\documentclass{article}

\usepackage[preprint]{neurips_2026}
\workshoptitle{Grounded and Faithful Vision-Language Models for Real-World Deployment (VLM4RWD)}

\usepackage[utf8]{inputenc}
\usepackage[T1]{fontenc}
\usepackage{hyperref}
\usepackage{url}
\usepackage{booktabs}
\usepackage{amsmath,amsfonts,amssymb}
\usepackage{microtype}
\usepackage{xcolor}
\usepackage{graphicx}
\usepackage{subcaption}

\newcommand{\E}{\mathcal{E}}
\newcommand{\Deltau}{\Delta}

\title{Conditional Visual Evidence Utility: State-Dependent Rank Reversals in Frozen Vision-Language Encoders}

\author{
  Yunxuan Fang \\
  Beihang University\\
  \texttt{micro@buaa.edu.cn} \\
  \And
  Xinhe Wang \\
  Beihang University \\
  \texttt{xinhe@buaa.edu.cn} \\
}

\begin{document}
\maketitle

\begin{abstract}
Static importance scores compress visual evidence into a single ranking, but the value of remaining evidence can change after one cue has been observed. We study this possibility in controlled compositional visual search, where color, shape, and texture evidence can be independently exposed and their conditional marginal utility measured across acquisition states. In a held-out confirmation on 800 scenes, frozen OpenCLIP and SigLIP exhibit robust state-dependent rank reversals that concentrate in candidate-overlap regimes designed to induce ordering changes, persist across two evidence-accumulation constructions and ten equivalent query wordings, and collapse to near-chance-scale behavior under query-scene derangement. A subsequent role-balanced follow-up on 1,200 scenes rotates the abstract roles of initially strong, redundancy-inducing, and comparator attributes; the positive-minus-negative reversal contrast remains positive across all 24 role-permutation, backbone, and evidence-mode cells, although residual attribute-identity effects remain. We further distinguish measured replanning opportunity from prospective predictability. Matched-first-action utility analyses show substantial opportunity to rerank remaining evidence, but lightweight predictors using posterior-based or acquired-embedding state representations do not establish a robust incremental advantage of acquired-state information over legal static controls on the role-balanced benchmark. Together, these results show that conditional visual evidence utility is reliably state dependent in this controlled setting, while separating the existence of changing utility from the stronger claim that those changes are prospectively predictable by a learned selector.
\end{abstract}

\section{Introduction}

Vision-language evidence use depends not only on whether relevant evidence is present, but also on how its value changes as other evidence becomes available. Most importance summaries are static: a region, token, feature, or modality receives a score that is interpreted without reference to the evidence already acquired. In this paper, we ask whether that abstraction is sufficient: \emph{Does visual evidence have a fixed importance ordering, or can its marginal utility change with the current evidence state?}

This distinction matters whenever evidence interacts. An attribute that is highly diagnostic at the start can become redundant once another attribute narrows the candidate set. Conversely, a previously weak attribute can become the most useful next observation. In such cases, the relevant quantity is not a static score $I(k)$ but the conditional marginal utility $\Deltau_k(S,q)$ of evidence type $k$ given acquired evidence $S$ and query $q$. If that ordering changes with $S$, a decision rule that commits to one ranking at the start can become stale. 
Sequential feature-acquisition methods already optimize information value conditioned on the current acquisition state. Here, we use conditional utility for a different purpose: as an empirical diagnostic of whether frozen vision-language encoders exhibit structured, state-dependent changes in the value of visual evidence.

We study this behavior with a controlled intervention setup. Each synthetic search scene contains 25 objects composed from color, shape, and texture. For every scene and query, we render isolated views that expose one evidence type at a time, evaluate every acquisition subset, and measure target utility with frozen vision-language encoders. The scene generator controls candidate overlap so that some regimes predict a change in the relative value of evidence while others do not. The controlled regimes specify when rank reversals should arise from candidate-overlap structure. We then test whether model-derived utilities reproduce these regime-dependent patterns, providing a behavioral test of state-dependent evidence valuation in frozen encoders.

Across two frozen backbones and two evidence-accumulation constructions, held-out confirmation shows the same qualitative pattern: robust reversals concentrate in the regimes that symbolically predict them, remain stable under changes in wording and evidence construction, and lose that structure when the query-scene relation is deliberately broken. A separate earlier frozen confirmation additionally shows that, within a fixed scene, the preferred next evidence changes with the semantic target query. We then repeat the structural test in a 1,200-scene role-balanced construction that rotates which attribute is initially strong, induces redundancy, and serves as the comparator. The positive-versus-negative reversal contrast remains positive for every role permutation under both backbones and evidence constructions, showing coverage beyond the original texture-centered assignments while retaining measurable attribute-identity asymmetries.

We also ask whether the change in ordering has a decision consequence. Since a global order also ignores scene/query differences already visible before acquisition, a prospectively specified adaptive-versus-global-fixed comparison was positive overall but did not isolate replanning. We therefore first use a clearly labeled \emph{post-confirmation exploratory} matched-first-action analysis: both policies make the same first acquisition and differ only in whether the remaining evidence is reranked afterwards. These measured-utility decisions retain positive utility when selected under one evidence mode, wording, or backbone and evaluated under another, establishing a substantial replanning opportunity in the saved utility landscape. We then test the stronger prospective question of whether acquired-state information predicts that second action beyond legal static information. Exact matched-static contexts remain substantially ambiguous, but stateful linear, nonlinear posterior-based, and embedding-based predictors do not establish a robust incremental advantage on the role-balanced benchmark. We therefore separate two claims that are easy to conflate: evidence utility can change with state even when the acquired observation does not support a reliably better learned next-action policy under the predictors tested here.

Our contributions are as follows:
\begin{itemize}
  \item a controlled intervention framework for measuring conditional utility over isolated, non-spatial visual evidence types;
  \item held-out evidence that state-dependent rank reversal is margin-robust, regime-specific, oracle-aligned, stable across evidence construction, wording, and backbone, and reproduced across all abstract attribute-role assignments in a role-balanced follow-up, with residual attribute-identity effects;
  \item a matched-first-action analysis that exposes substantial replanning opportunity, together with prospective negative results showing that this opportunity does not automatically translate into incremental history-based predictability.
\end{itemize}

\section{Related work}

\paragraph{Sequential evidence acquisition and active vision.}
Sequential acquisition has a long history in dynamic feature selection, where the next observation is chosen from the current acquisition state, including conditional-information formulations such as DIME \citep{covert2023dynamic,gadgil2024dime}. Active-vision systems extend this idea to visual evidence, adapting spatial location, scale, frame, or temporal interval during inference \citep{wu2023vstar,liu2026fovea,wang2025avp}. Our setting differs in the object being measured: with spatial correspondence held fixed, we examine how the marginal utility ordering over distinct \emph{types} of visual evidence changes as evidence accumulates.

\paragraph{History-conditioned multimodal reasoning.}
Recent multimodal systems increasingly use accumulated observations or memory to guide subsequent retrieval, inspection, or perceptual refinement \citep{wang2025avp,yamao2026qvic}. For example, QViC-MF feeds relevant past visual context back into question-guided compression of the current clip, allowing previously stored evidence to influence what ongoing perception preserves. These systems operationalize history-conditioned decision making at the policy or perception level. We instead study the underlying utility structure directly, using controlled evidence states to measure when the preferred next observation changes and whether those changes follow an independently specified candidate-overlap structure.

\paragraph{Visual-token routing and static importance.}
Token-pruning methods such as FastV and SparseVLM rank or route spatial visual tokens to reduce computation \citep{chen2024fastv,zhang2025sparsevlm}. Related diagnostic work has shown that static importance estimates can be fragile, with heuristic or text-guided scores sometimes failing to outperform simple baselines or suffering from cross-modal misalignment \citep{xu2026rethinking,wen2025token}. Our analysis addresses a complementary issue: whether evidence utility itself is state dependent, such that a single fixed ranking cannot describe all acquisition states. Instead of treating an attention or saliency score as evidence utility, we directly vary the available color, shape, and texture evidence and measure the resulting target utility across acquisition states.

\section{Formulation and Setup}
\label{sec:method}

\subsection{Problem formulation}

Let $q$ denote a unique target among $N=25$ candidate objects. The evidence universe is $\E=\{c,s,t\}$ for color, shape, and texture, and $S\subseteq\E$ denotes acquired evidence. For each rendered evidence state, we crop the same 25 candidate cells, encode each crop with the frozen image encoder, encode the text query once, and use the model's image-text similarity logits as candidate scores. A softmax over the 25 candidate logits defines the target probability. We then define
\begin{equation}
  U(S,q)=\log p(y^*\mid S,q), \qquad
  \Deltau_k(S,q)=U(S\cup\{k\},q)-U(S,q).
  \label{eq:utility}
\end{equation}
Thus $\Deltau_k$ is both the increase in target log probability and the decrease in target negative log likelihood.

For two available evidence groups $i,j\notin S$, write $d_{ij}(S,q)=\Deltau_i(S,q)-\Deltau_j(S,q)$. A state-dependent rank reversal occurs between the empty state and an acquired state $H$ when $d_{ij}(\varnothing,q)d_{ij}(H,q)<0$. Our robust definition additionally requires
\begin{equation}
 |d_{ij}(\varnothing,q)|>\epsilon
 \quad\text{and}\quad
 |d_{ij}(H,q)|>\epsilon,
 \label{eq:robust}
\end{equation}
with prospectively fixed primary threshold $\epsilon=0.05$. A sample reverses if any of its three attribute pairs satisfies Eq.~\ref{eq:robust} using the third attribute as history. We emphasize this margin-based statistic because unthresholded any-pair rates are permissive.

\subsection{Controlled compositional setup}

Each scene contains a $5\times5$ grid of objects sampled from six colors, six shapes, and four textures. The query is a conjunction such as ``the red striped triangle''. We generate isolated views in which only the named attribute varies across candidate locations while the other two attributes are rendered canonically. Consequently, each intervention preserves spatial correspondence but controls which information type distinguishes candidates. Full construction and sanity checks are in Appendix~\ref{app:setup}.

Four regimes manipulate symbolic candidate overlap. In \emph{color-initially-strongest-then-redundant} and \emph{shape-initially-strongest-then-redundant}, the initially diagnostic attribute becomes redundant after another attribute is acquired. These are oracle-positive. \emph{Matched singleton discriminativeness} and \emph{weak complementary attributes} are oracle-negative under the non-tied definition. The symbolic oracle is a design diagnostic computed only from candidate-set cardinalities, without using OpenCLIP or SigLIP outputs. The generator therefore specifies where an ordering change should be structurally expected, but the model is free to agree or disagree. Agreement with the oracle does not imply that a model implements its symbolic rule internally.

Let $R_+$ and $R_-$ denote sample-level robust reversal rates in oracle-positive and oracle-negative regimes. We summarize regime specificity with
\begin{equation}
  \Gamma_{\mathrm{rev}} = R_+ - R_-,
  \label{eq:regime-contrast}
\end{equation}
reported in percentage points. A large positive $\Gamma_{\mathrm{rev}}$ indicates that reversals follow the intended candidate-overlap manipulation rather than appearing uniformly across scenes.

\begin{figure}[t]
  \centering
  \begin{subfigure}[b]{0.22\linewidth}\centering
    \includegraphics[width=\linewidth]{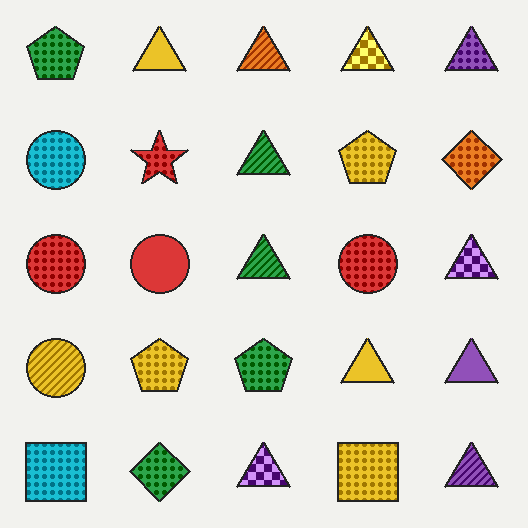}
    \caption{Full scene}
  \end{subfigure}\hfill
  \begin{subfigure}[b]{0.22\linewidth}\centering
    \includegraphics[width=\linewidth]{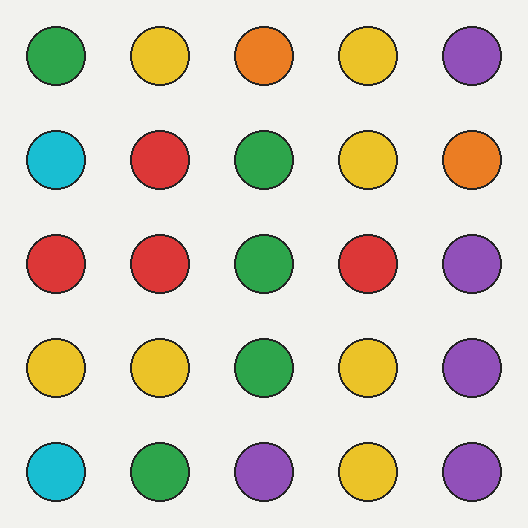}
    \caption{Color view}
  \end{subfigure}\hfill
  \begin{subfigure}[b]{0.22\linewidth}\centering
    \includegraphics[width=\linewidth]{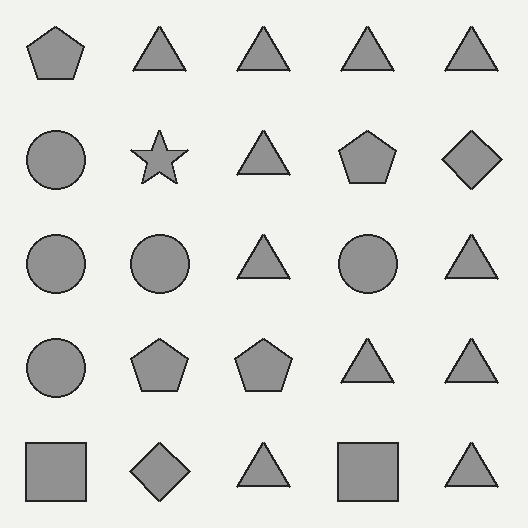}
    \caption{Shape view}
  \end{subfigure}\hfill
  \begin{subfigure}[b]{0.22\linewidth}\centering
    \includegraphics[width=\linewidth]{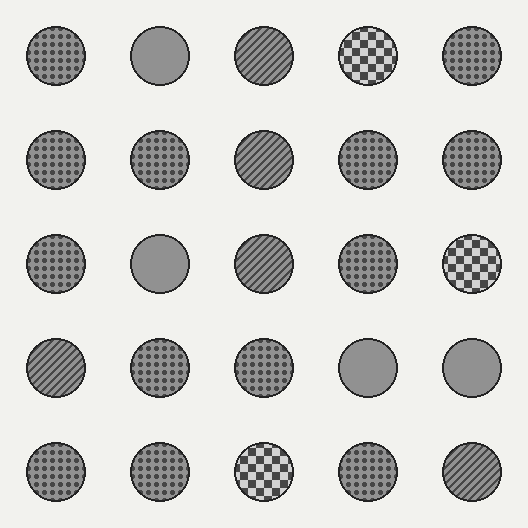}
    \caption{Texture view}
  \end{subfigure}
  \vspace{2mm}
  \fbox{\parbox{0.92\linewidth}{\centering
  Empty state: $\Deltau_c>\Deltau_t$ \quad $\longrightarrow$ acquire shape \quad
  Conditioned state: $\Deltau_c(\{s\})<\Deltau_t(\{s\})$.}}
  \caption{Controlled evidence intervention and rank reversal. Isolated views preserve candidate locations while exposing one attribute type. The schematic ordering is illustrative, reported reversals are computed from frozen-model utilities.}
  \label{fig:concept}
\end{figure}

\subsection{Evaluation protocol and controls}

\paragraph{Frozen confirmation.}
The held-out split contains 800 new scenes, exactly 200 per regime, disjoint from all exploratory and development scenes. The analysis protocol, scene manifests, query templates, controls, and hashes were frozen before inference on this split. We evaluate OpenCLIP ViT-B/32 (\texttt{laion2B-s34B-b79K}) \citep{ilharco2021openclip} and SigLIP base patch16-224 \citep{zhai2023siglip}, both remain frozen, with no calibration or fine-tuning.

\paragraph{Evidence accumulation.}
In \emph{additive logits}, we sum candidate logits from separately encoded singleton views. The empty state uses an all-zero candidate-logit vector, yielding a uniform distribution over the 25 candidates. We use the same empty-state baseline for \emph{direct render}; non-empty accumulated subsets are jointly rendered and encoded. Thus the two constructions share the same initial state and singleton evidence, while direct rendering additionally permits interactions within combined visual inputs. Agreement across them tests whether the reversal depends on one particular evidence construction.

\paragraph{Statistics.}
Primary results use template T1 (``the \{color\} \{texture\} \{shape\}'') and $\epsilon=0.05$. We report $\Gamma_{\mathrm{rev}}$ together with sample-level oracle balanced accuracy. All 95\% intervals use 10,000 percentile bootstrap repetitions with scene as the resampling cluster, regime contrasts resample within positive and negative strata. Secondary analyses cover $\epsilon\in\{0,.1,.25,.5\}$, ten equivalent wordings, cross-mode agreement, controls, and selectivity.

\paragraph{Setup validity and controls.}
Each isolated view must decode its intended attribute above a frozen threshold. Both backbones obtain 100\% color, 100\% shape, and 74.86\% texture accuracy. The prespecified removed-attribute diagnostic does not trigger its predefined strong-leakage flag. This result rules out the prespecified strong-leakage condition but does not establish that removed attributes are at chance (Appendix~\ref{app:robustness}). 
A query-scene derangement preserves recipient images and target indices but replaces the query with an absent conjunction from another scene. A uniform-target implementation sanity check is reported in the appendix.

\paragraph{Role-balanced and prospective-predictor follow-ups.}
After the 800-scene confirmation, we construct a separate 1,200-scene role-balanced dataset that rotates the abstract roles of initially strong, redundancy-inducing, and comparator attributes across all six assignments of color, shape, and texture. We evaluate the same robust-reversal criterion under both frozen backbones and both evidence modes. We additionally test prospective second-action prediction under a fixed scene-level split. Predictor inputs are restricted to legal current-state and query information; regime labels, role labels, future utilities, and unacquired evidence are excluded. The learned-policy analyses are treated as exploratory predictability tests rather than extensions of the frozen confirmation.

\section{Results}

\subsection{State-dependent utility reversals are regime-specific}

\begin{figure}[t]
\centering
\includegraphics[width=0.9\linewidth]{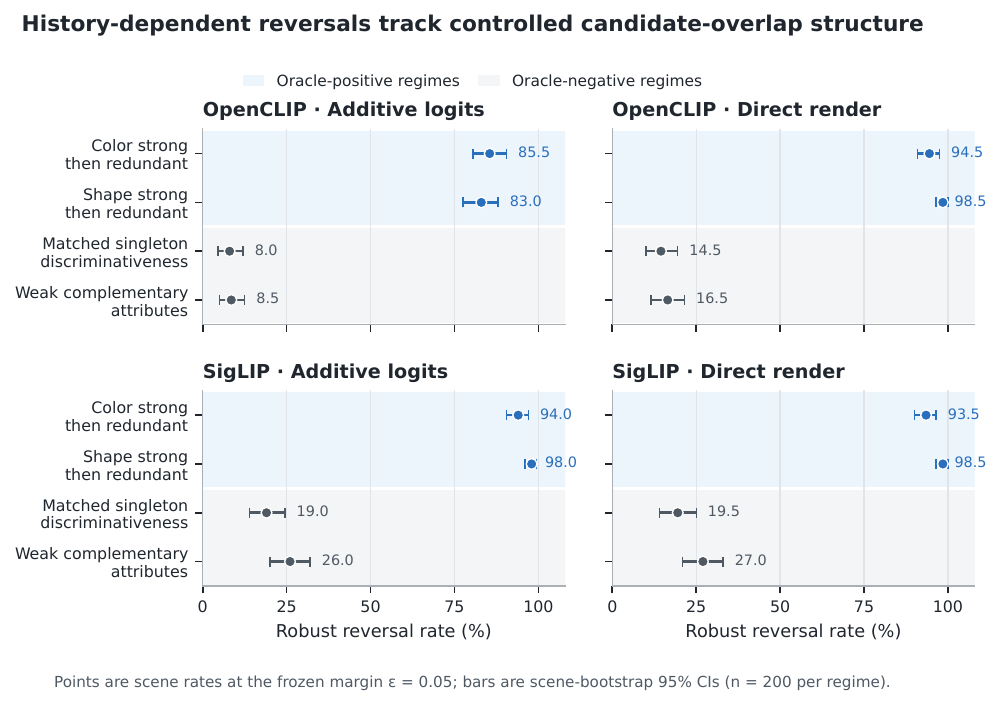}
\caption{Confirmed state-dependent rank reversal across backbones and evidence constructions. Robust reversal rates are substantially higher in oracle-positive than oracle-negative regimes, error bars show scene-bootstrap 95\% confidence intervals.}
\label{fig:confirmed}
\end{figure}

Table~\ref{tab:headline} and Fig.~\ref{fig:confirmed} show the confirmatory result. Across all four backbone/mode cells, robust reversals are concentrated in oracle-positive regimes, not uniformly across scenes. The resulting regime contrast $\Gamma_{\mathrm{rev}}$ is 72.7--81.0 percentage points, with every 95\% interval excluding zero, while oracle balanced accuracy remains high across both backbones and evidence constructions. The key result is therefore not merely that reversals occur, but that their occurrence follows the candidate-overlap manipulation designed to change the relative value of evidence. Pair-level analysis further localizes the effect: robust reversals concentrate in the two attribute comparisons for which the symbolic construction predicts ordering changes, whereas color-versus-shape reversals remain uncommon (Appendix~\ref{app:robustness}).

\begin{table}[t]
\centering
\small
\caption{Frozen confirmation at $\epsilon=0.05$ on 800 scenes. RR is sample-level robust reversal rate, BA is sample-level oracle balanced accuracy, CIs are scene-bootstrap intervals.}
\label{tab:headline}
\resizebox{\linewidth}{!}{%
\begin{tabular}{llrrrr}
\toprule
Backbone & Evidence mode & RR positive (\%) & RR negative (\%) & Difference (pp) [95\% CI] & Oracle BA (\%) \\
\midrule
OpenCLIP & additive & 84.2 & 8.2 & 76.0 [71.5, 80.2] & 88.0 \\
OpenCLIP & direct render & 96.5 & 15.5 & 81.0 [77.0, 85.0] & 90.5 \\
SigLIP & additive & 96.0 & 22.5 & 73.5 [69.0, 78.0] & 86.8 \\
SigLIP & direct render & 96.0 & 23.2 & 72.7 [68.0, 77.2] & 86.4 \\
\bottomrule
\end{tabular}}
\end{table}

\subsection{Reversals are robust to evidence construction and sensitive to query semantics}

Two complementary checks test whether the result is tied to one evidence construction. Reversal labels, conditional utilities, and preferred next actions agree strongly between additive and jointly rendered accumulation for both backbones. Direct rendering changes some individual decisions, as expected from visual interactions, but preserves the same regime-level structure. The reversal phenomenon is therefore not an artifact of simply summing singleton logits.

Every frozen wording template in the 800-scene confirmation preserves a positive $\Gamma_{\mathrm{rev}}$, showing that the effect is not tied to one surface form of the query. Surface-form robustness does not by itself establish semantic query dependence. 
A separate earlier frozen confirmation on 400 scenes varied the semantic target while holding the scene fixed. Across 20,670 matched scene-state-target observations, query-specific symbolic-oracle action agreement was 92.06--94.84\%, versus 61.44\% for a query-independent per-scene/state modal-oracle baseline. This corresponds to an improvement of 30.62--33.40 percentage points across both backbones and evidence modes; all scene-bootstrap 95\% confidence intervals excluded zero. We report this as prior frozen confirmation; full cells and switch-direction statistics are in Appendix~\ref{app:semantic-query}.

Conversely, in the current confirmation, query-scene derangement destroys localization and returns oracle alignment and the regime contrast to approximately chance-scale behavior. As recipient images and target indices are preserved, this control shows that the regime-level structure depends on the intended query-scene relation rather than recipient layout or target-index statistics alone. Together, the wording, semantic-target, and derangement controls support query-sensitive conditional evidence use rather than dependence on a particular prompt template. Exact agreement statistics, threshold sensitivities, and current-run control values are reported in Appendix~\ref{app:robustness}.

\subsection{Role-balanced follow-up covers all abstract role assignments}
\label{sec:role-balanced}

The original confirmation deliberately exposed a limitation: both oracle-positive regimes used texture as one side of the reversing comparison, so the observed effect could not distinguish a general role-based mechanism from an attribute-specific construction. The role-balanced follow-up rotates color, shape, and texture through all six assignments of initially strong, redundancy-inducing, and comparator roles. At $\epsilon=0.05$, the intended-pair positive-minus-negative reversal contrast remains large under every backbone and evidence mode (Table~\ref{tab:role-balanced}). More strongly, all 24 role-permutation $\times$ backbone $\times$ mode contrasts have 95\% intervals above zero; the smallest lower bound is 46.4 percentage points. The state-dependent ordering effect is therefore not restricted to texture serving as the comparator.

\begin{table}[t]
\centering
\small
\caption{Role-balanced follow-up at $\epsilon=0.05$ on 1,200 scenes. Rates are intended-pair robust reversals; CIs are scene-bootstrap intervals for the positive-minus-negative contrast.}
\label{tab:role-balanced}
\begin{tabular}{llrrr}
\toprule
Backbone & Evidence mode & Positive (\%) & Negative (\%) & Difference (pp) [95\% CI] \\
\midrule
OpenCLIP & additive & 68.37 & 0.88 & 67.48 [64.02, 70.92] \\
OpenCLIP & direct render & 72.15 & 9.00 & 63.15 [59.65, 66.65] \\
SigLIP & additive & 71.82 & 0.95 & 70.87 [68.63, 73.08] \\
SigLIP & direct render & 76.95 & 9.08 & 67.87 [65.18, 70.50] \\
\bottomrule
\end{tabular}
\end{table}

Role coverage is not the same as permutation invariance. Seven unadjusted descriptive attribute-identity contrasts in positive regimes exclude zero. The clearest recurring pattern is a lower reversal rate when the initially strong attribute is color rather than texture, with differences of $-12.15$ pp for OpenCLIP additive, $-12.85$ pp for OpenCLIP direct, and $-7.20$ pp for SigLIP additive. We therefore interpret the follow-up as evidence for coverage across all abstract role assignments with residual attribute-identity effects, not as evidence that attribute identity is irrelevant.

\subsection{Exploratory value of replanning}
\label{sec:replan}

The prospectively specified adaptive-versus-global-fixed comparison showed an overall utility advantage, but no planned regime interaction (Appendix~\ref{app:planned-adaptive}). To isolate updating after evidence acquisition from differences already present at the initial state, we performed a post-confirmation exploratory analysis in which the competing policies are constrained to take the same first action.

Both policies choose
\begin{equation}
 a_1=\arg\max_{k\in\E}\Deltau_k(\varnothing,q).
\end{equation}
Commit-once then follows the second-ranked remaining group from the empty-state order. Replan instead selects
\begin{equation}
 a_2=\arg\max_{k\ne a_1}\Deltau_k(\{a_1\},q).
\end{equation}
The two policies have identical costs, first states, and final all-evidence states, only the second acquisition differs.

\begin{figure}[t]
\centering
\includegraphics[width=0.9\linewidth]{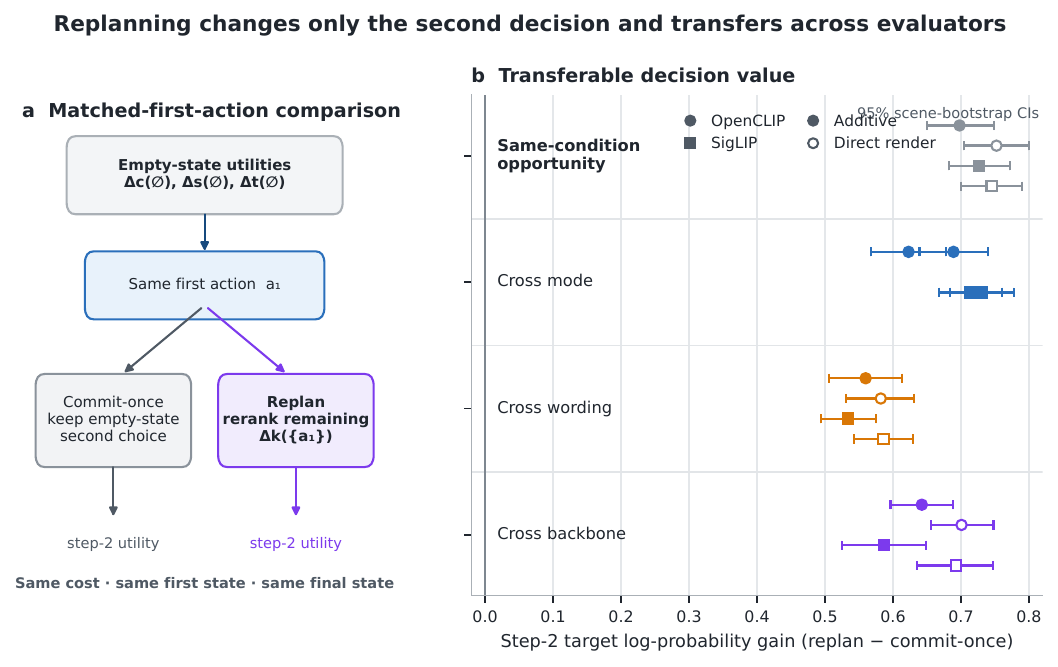}
\caption{Post-confirmation exploratory value of replanning. Commit-once and replan share the same first action and differ only at step 2. Cross-condition gains evaluate actions selected under one evidence mode, wording, or backbone with a different evaluator, error bars show scene-bootstrap 95\% confidence intervals.}
\label{fig:replan}
\end{figure}

Same-condition gains provide an opportunity upper bound because the replan action is selected using the evaluator's own counterfactual utilities. Cross-condition evaluation provides a stricter test: the second action is selected under one evidence mode, wording, or backbone and scored under another. All 12 cross-condition scene-bootstrap intervals remain above zero (Table~\ref{tab:replan} and Fig.~\ref{fig:replan}), with substantially larger gains in oracle-positive regimes than in oracle-negative regimes. This supports the presence of decision-relevant structure that persists across evaluator changes in this setup, beyond the same-condition maximization effect. Because these comparisons preserve the same scenes and semantic targets, they assess robustness to evaluator changes, not out-of-distribution generalization.

\begin{table}[t]
\centering
\small
\caption{Post-confirmation exploratory matched-first-action results. Ranges cover all directional backbone/mode cells, every individual 95\% CI is positive. Full cells are in Appendix~\ref{app:replan}.}
\label{tab:replan}
\begin{tabular}{lccc}
\toprule
Evaluation & Cells & Step-2 utility gain & Action agreement (\%) \\
\midrule
Cross mode & 4 & 0.623--0.730 & 88.5--94.5 \\
T1 $\rightarrow$ held-out wording & 4 & 0.534--0.586 & 88.4--92.1 \\
Cross backbone & 4 & 0.587--0.701 & 85.4--89.9 \\
\bottomrule
\end{tabular}
\end{table}

\subsection{Prospective predictors do not establish incremental history value}
\label{sec:predictability}

The measured-utility analysis above selects the second action using counterfactual utilities and therefore quantifies opportunity rather than a deployable policy. We next ask whether the acquired observation contributes predictive information beyond legal static context. The role-balanced benchmark is not statically determined under the exact legal key $(a_1, q_{\mathrm{color}}, q_{\mathrm{shape}}, q_{\mathrm{texture}}, \text{template})$: empirical conditional entropy is 0.631--0.722 bits across the four backbone/mode conditions, 71.8--79.7\% of rows lie in ambiguous static strata, and each condition contains 976--1,043 opposite-label matched pairs. A train-only modal static lookup remains far below the measured-utility oracle. Thus a null stateful result cannot be explained by the second action being trivially encoded in the allowed static metadata.

Nevertheless, increasingly expressive state representations do not yield a stable prospective advantage on the role-balanced data. Linear stateful predictors using the acquired posterior do not improve reliably over matched no-history controls. Permutation-invariant posterior representations with nonlinear tree and MLP predictors likewise yield no significant matched-ambiguity cross-entropy improvement in any of the four role-balanced conditions. Finally, a small permutation-invariant set encoder over the acquired frozen visual embeddings also fails the predefined method stop rule (Table~\ref{tab:embedding-predictor}). Only one of four matched-ambiguity cross-entropy point estimates is positive; the scene-pooled stateful-minus-no-history improvement is $-0.0312$ bits with 95\% CI $[-0.0728, 0.0084]$. The pooled matched-subset utility difference is $0.0217$ with CI $[-0.0527, 0.0953]$, and all-test utility is likewise indistinguishable from zero.

\begin{table}[t]
\centering
\small
\caption{Final embedding-state replanner on the role-balanced benchmark. Positive CE improvement means lower held-out cross-entropy for stateful than matched no-history.}
\label{tab:embedding-predictor}
\begin{tabular}{llr}
\toprule
Backbone & Evidence mode & Matched CE improvement, bits [95\% CI] \\
\midrule
OpenCLIP & additive & $-0.0302$ [$-0.0838$, 0.0252] \\
OpenCLIP & direct render & $-0.1044$ [$-0.2106$, $-0.0028$] \\
SigLIP & additive & 0.0136 [$-0.0068$, 0.0335] \\
SigLIP & direct render & $-0.0002$ [$-0.0088$, 0.0081] \\
\bottomrule
\end{tabular}
\end{table}

These results separate the behavioral phenomenon from its prospective exploitation. The benchmark contains genuine state-dependent changes in measured utility and substantial matched-static ambiguity, but the acquired-state representations tested here do not provide a robust incremental signal for predicting the utility-optimal second action. We therefore stop predictor architecture search rather than interpreting measured replanning opportunity as evidence for a learned history-conditioned router.

\section{Discussion and Future Directions}

\paragraph{Grounded evaluation should test importance conditionally.}
The main implication is methodological. A single evidence ordering assigned to an entire decision episode can represent at most one acquisition state; it cannot express state-specific reorderings. Evaluations of grounding, attribution, or evidence selection should therefore report how evidence value changes across matched acquisition states, for example through conditional marginal-utility curves, rank-stability statistics, or reversal rates. The controlled positive/negative regimes in our setup also show why aggregate reversal counts are insufficient: a useful evaluation should ask whether ordering changes occur where the evidence structure predicts an interaction, rather than merely whether any pair flips somewhere.

\paragraph{Changing utility does not guarantee a predictable replanning policy.}
For systems that acquire or inspect evidence sequentially, a changing utility ordering creates an opportunity for replanning, but opportunity alone does not imply that the useful update can be predicted from the acquired observation. Our matched-first-action measured-utility analysis establishes the former: after the first acquisition, choosing the utility-optimal remaining action can improve the intermediate state, and those decisions retain value across evaluator changes. The prospective follow-ups test the stronger claim. Despite substantial matched-static ambiguity, posterior-based and embedding-based stateful predictors do not obtain a robust incremental advantage over legal no-history controls on the role-balanced benchmark. Conditional utility structure should therefore be distinguished from policy learnability rather than treated as sufficient evidence for a deployable adaptive selector.

\paragraph{Scope and extensions.}
The controlled construction remains intentionally narrow, but the role-balanced follow-up removes one important ambiguity in the original design. State-dependent reversal appears under all six assignments of color, shape, and texture to the abstract roles of initially strong, redundancy-inducing, and comparator evidence. At the same time, residual attribute-identity effects remain, so the result supports role coverage rather than full permutation invariance.

More broadly, color, shape, and texture are controlled input interventions, so the present results characterize behavioral evidence use without implying an internal processing order, separable latent concepts, or independently executable compute units. The negative prospective-predictor results further caution against assuming that a measured conditional-utility effect automatically yields a usable history-conditioned policy. A stronger test of practical adaptive acquisition may require richer evidence sources whose acquired content exposes information relevant to the value of later sources, such as OCR, object state, relations, motion, or audio, together with independently executable evidence groups and real acquisition or compute costs. Those extensions are outside the claims of the present study.

\section{Conclusion}

Visual evidence importance is not necessarily a property of the evidence alone. In our controlled setup, it depends on both the acquired evidence state and the semantic target query. Across two frozen vision-language encoders, robust rank reversals follow the candidate-overlap regimes designed to change evidence utility, persist across evidence constructions and query wordings, and disappear when the query-scene relation is broken. A separate prior frozen confirmation shows that preferred next evidence also changes with the semantic target within a fixed scene, and a 1,200-scene role-balanced follow-up shows that the reversal mechanism appears across all abstract attribute-role assignments while retaining residual attribute-identity effects. Matched-first-action analysis further reveals substantial measured replanning opportunity across evaluator changes, but prospective posterior-based and embedding-based predictors do not establish a robust incremental value of acquired-state information on the role-balanced benchmark. The resulting distinction is central: state-dependent utility is a reproducible behavioral property in this assay, but it does not by itself imply a prospectively learnable history-conditioned policy. Conditional evidence utility therefore provides a target for evaluating how evidence value changes across states while also defining a boundary that future adaptive acquisition methods must independently overcome.

\bibliographystyle{plainnat}
\bibliography{references}

@inproceedings{gadgil2024dime,
  title={Estimating Conditional Mutual Information for Dynamic Feature Selection},
  author={Gadgil, Soham and Covert, Ian and Lee, Su-In},
  booktitle={International Conference on Learning Representations},
  year={2024}
}

@inproceedings{covert2023dynamic,
  title={Learning to Maximize Mutual Information for Dynamic Feature Selection},
  author={Covert, Ian Connick and Qiu, Wei and Lu, Mingyu and Kim, Na Yoon and White, Nathan J. and Lee, Su-In},
  booktitle={International Conference on Machine Learning},
  year={2023}
}

@article{wu2023vstar,
  title={V*: Guided Visual Search as a Core Mechanism in Multimodal LLMs},
  author={Wu, Penghao and Xie, Saining},
  journal={arXiv preprint arXiv:2312.14135},
  year={2023}
}

@article{liu2026fovea,
  title={The Perceptual Bandwidth Bottleneck in Vision-Language Models},
  author={Liu, Yiming and others},
  journal={arXiv preprint arXiv:2605.01345},
  year={2026}
}

@article{wang2025avp,
  title={Active Video Perception: Iterative Evidence Seeking for Agentic Long Video Understanding},
  author={Wang, Ziyang and Zhou, Honglu and Wang, Shijie and Li, Junnan and Xiong, Caiming and Savarese, Silvio and Bansal, Mohit and Ryoo, Michael S. and Niebles, Juan Carlos},
  journal={arXiv preprint arXiv:2512.05774},
  year={2025}
}

@article{chen2024fastv,
  title={An Image is Worth 1/2 Tokens After Layer 2: Plug-and-Play Inference Acceleration for Large Vision-Language Models},
  author={Chen, Liang and others},
  journal={arXiv preprint arXiv:2403.06764},
  year={2024}
}

@article{zhang2025sparsevlm,
  title={SparseVLM: Visual Token Sparsification for Efficient Vision-Language Model Inference},
  author={Zhang, Yuan and others},
  journal={arXiv preprint arXiv:2410.04417},
  year={2025}
}

@article{ilharco2021openclip,
  title={OpenCLIP},
  author={Ilharco, Gabriel and Wortsman, Mitchell and Wightman, Ross and Gordon, Cade and Carlini, Nicholas and Taori, Rohan and Dave, Achal and Shankar, Vaishaal and Namkoong, Hongseok and Miller, John and Hajishirzi, Hannaneh and Farhadi, Ali and Schmidt, Ludwig},
  journal={Zenodo},
  year={2021},
  doi={10.5281/zenodo.5143773}
}

@inproceedings{zhai2023siglip,
  title={Sigmoid Loss for Language Image Pre-Training},
  author={Zhai, Xiaohua and Mustafa, Basil and Kolesnikov, Alexander and Beyer, Lucas},
  booktitle={International Conference on Computer Vision},
  year={2023}
}

@inproceedings{wen2025token,
  title={Token pruning in multimodal large language models: Are we solving the right problem?},
  author={Wen, Zichen and Gao, Yifeng and Li, Weijia and He, Conghui and Zhang, Linfeng},
  booktitle={Findings of the Association for Computational Linguistics: ACL 2025},
  pages={15537--15549},
  year={2025}
}

@inproceedings{xu2026rethinking,
  title={Rethinking visual token reduction in lvlms under cross-modal misalignment},
  author={Xu, Rui and Wang, Yunke and Luo, Yong and Du, Bo},
  booktitle={Proceedings of the AAAI Conference on Artificial Intelligence},
  volume={40},
  number={32},
  pages={27323--27331},
  year={2026}
}

@inproceedings{yamao2026qvic,
  title={Question-guided Visual Compression with Memory Feedback for Long-Term Video Understanding},
  author={Yamao, Sosuke and Miyahara, Natsuki and Qi, Yuankai and Takeuchi, Shun},
  booktitle={Proceedings of the IEEE/CVF Conference on Computer Vision and Pattern Recognition},
  year={2026}
}

\appendix

\section{Setup construction and reproducibility}
\label{app:setup}

The held-out confirmation split uses base seed 20280823 and deterministic sample seeds. It contains 200 scenes in each of four regimes. Every target conjunction is unique in its scene. Ten query templates express the same conjunction; T1 is primary and T2--T10 are repeated wording measurements. All scene metadata, exact symbolic transitions, query strings, controls, configuration, and SHA256 hashes were frozen before model inference. Raw saved outputs contain 384,000 transition rows and 3.2 million candidate-logit rows per backbone. Analysis loads no model.

The two evidence modes produce the same subset lattice. For each state rendering, the 25 spatially corresponding candidate cells are cropped and independently encoded by the frozen image encoder; the text query is encoded once, and image-text similarity logits provide the 25 candidate scores. Additive logits sum scores from separately encoded singleton views. Both modes use the same all-zero candidate-logit empty state. For non-empty subsets, direct rendering constructs the accumulated evidence jointly before cropping and encoding. In both cases, $U(S,q)$ is target log probability after a softmax over the same 25 candidate locations. Frozen-model assertions verify evaluation mode, inference mode, and no trainable parameters. The frozen confirmation bootstrap uses 10,000 repeats and seed 20280826.

\section{Robustness and controls}
\label{app:robustness}

The primary threshold was chosen in an earlier exploratory forensic analysis and fixed before the 800-scene run. Sensitivity thresholds $0,.1,.25,.5$ were also frozen; none may replace the primary endpoint. The sample-level primary reversal rates (without regime aggregation) are 46.3\%, 56.0\%, 59.2\%, and 59.6\% for OpenCLIP-additive, OpenCLIP-direct, SigLIP-additive, and SigLIP-direct. Pair-level rates are 29.8\%, 36.0\%, 36.7\%, and 41.0\%. We report these only as scale summaries because regime specificity and oracle agreement are more diagnostic than a permissive any-pair aggregate.

\begin{table}[h]
\centering
\small
\caption{Pair-specific robust reversal rates at $\epsilon=0.05$ on the 800-scene confirmation. Each cell has $n=800$ scenes; intervals use 10,000 scene-bootstrap repeats. The conditioning attribute is the acquired evidence state used to compare the remaining pair.}
\label{tab:pair-breakdown}
\resizebox{\linewidth}{!}{%
\begin{tabular}{lrrrr}
\toprule
Comparison & OpenCLIP Add & OpenCLIP Direct & SigLIP Add & SigLIP Direct \\
\midrule
color $\leftrightarrow$ shape, after texture & 7.0 [5.2, 8.8] & 15.4 [12.9, 18.0] & 5.8 [4.2, 7.5] & 17.0 [14.4, 19.6] \\
color $\leftrightarrow$ texture, after shape & 41.0 [37.6, 44.5] & 45.6 [42.1, 49.1] & 52.2 [48.8, 55.6] & 52.8 [49.4, 56.1] \\
shape $\leftrightarrow$ texture, after color & 41.5 [38.1, 44.9] & 47.0 [43.5, 50.4] & 52.0 [48.5, 55.2] & 53.2 [49.8, 56.8] \\
\bottomrule
\end{tabular}}
\end{table}

The pair breakdown matches the symbolic construction at a finer granularity. The oracle predicts no color-shape reversal for any of the 800 scenes, whereas in the 400 oracle-positive scenes it predicts reversals for both texture-containing comparisons. Correspondingly, color-shape reversals are uncommon while both texture-containing pairs replicate across backbones and evidence modes. Because texture is the common comparator in both positive constructions, this table localizes the present effect but does not establish attribute-role invariance.

The uniform-target control is an implementation sanity check rather than evidence for the behavioral claim. It contains 8,000 target draws per backbone and yields a 3.6125\% true-target draw rate against the exact 4\% reference. Deranged queries are deterministically matched across scenes without fixed points, and their conjunction is absent from the recipient scene. For OpenCLIP, deranged oracle BA is 50.25\%/52.63\% (additive/direct), versus 88.0\%/90.5\% normally. For SigLIP it is 48.88\%/51.0\%, versus 86.75\%/86.38\%. Deranged positive-minus-negative reversal contrasts are 0.5, 5.25, $-2.25$, and 2.0 percentage points, respectively.

The 3-by-3 selectivity setup evaluates intended and removed attributes without a trained probe. Both backbones score 100\% on isolated color and shape and 74.86\% on isolated texture. The prespecified diagnostic flags strong leakage only when a removed-attribute balanced-accuracy lower confidence bound exceeds chance by 0.10; no cell triggers that flag. This criterion is a one-sided screen for detectable strong leakage, not an equivalence test, so its failure to trigger should not be read as evidence that removed-attribute decoding is exactly at chance.

\section{Prior frozen semantic-target confirmation}
\label{app:semantic-query}

A separate earlier frozen confirmation tested semantic query dependence by changing the target query while holding each scene fixed. This experiment used 400 scenes and 20,670 matched observations and was specified prospectively as primary criterion P3 in its own frozen protocol. It is not part of the later 800-scene confirmation reported as the main result.

Evaluation includes only matched scene/state units containing at least two semantic target queries with different non-tied oracle-optimal actions; two-attribute states with only one available action are excluded.

\begin{table}[h]
\centering
\small
\caption{Prior frozen semantic-target confirmation. Agreement measures whether the model's marginal-utility-maximizing next evidence type matches the query-specific non-tied symbolic-oracle action. The query-independent baseline assigns one modal oracle action to each matched scene/state without using the current target query. Improvements are percentage points; intervals use scene-bootstrap resampling.}
\label{tab:semantic-query}
\resizebox{\linewidth}{!}{%
\begin{tabular}{llrrr}
\toprule
Backbone & Evidence mode & Query-specific oracle-action agreement (\%) & Query-independent modal baseline (\%) & Improvement (pp) [95\% CI] \\
\midrule
OpenCLIP & additive & 92.06 & 61.44 & 30.62 [29.97, 31.28] \\
OpenCLIP & direct render & 92.75 & 61.44 & 31.31 [30.65, 31.97] \\
SigLIP & additive & 94.74 & 61.44 & 33.30 [32.66, 33.95] \\
SigLIP & direct render & 94.84 & 61.44 & 33.40 [32.78, 34.01] \\
\bottomrule
\end{tabular}}
\end{table}

Among matched query pairs for which the symbolic oracle requires the preferred next action to change, model switch rates are 85.6--90.4\%, and agreement with the oracle's switch direction is 82.0--87.1\% across backbone/mode cells. These results directly test semantic-target dependence rather than paraphrase robustness: the same visual scene can induce a different preferred next evidence type when the target query changes.

\section{Planned adaptive comparison and its negative interaction}
\label{app:planned-adaptive}

The frozen protocol enumerated all six global fixed orders and compared them with a sample-wise measured-utility adaptive order. The best global fixed order by mean target-log-probability AUC was color-texture-shape in all four backbone/mode cells. Adaptive-minus-best-fixed AUC was positive: 0.562 [0.504, 0.621], 0.447 [0.394, 0.501], 0.441 [0.400, 0.482], and 0.447 [0.401, 0.491]. However, the planned mechanism-relevant interaction, whether this advantage is larger in oracle-positive regimes, was $-0.075$, $-0.072$, 0.003, and $-0.002$. We retain this negative result. A global order cannot adapt even to information already present at the empty state, so its overall disadvantage does not isolate the value of updating after acquisition.

\section{Full exploratory replanning transfers}
\label{app:replan}

This analysis was specified after the confirmation results were inspected, uses only saved transitions, and was not part of the frozen confirmation protocol. Bootstrap resampling uses 10,000 repeats, seed 20280827, and scenes as clusters. Table~\ref{tab:replanfull} gives all cross-condition cells.

\begin{table}[h]
\centering
\small
\caption{Exploratory cross-condition step-2 target-log-probability gain, replan minus commit-once. OC and SL denote OpenCLIP and SigLIP; Add and Dir denote additive and direct-render modes.}
\label{tab:replanfull}
\begin{tabular}{lll}
\toprule
Transfer & Direction & Gain [95\% CI] \\
\midrule
Mode & OC Add $\rightarrow$ Dir & 0.689 [0.639, 0.740] \\
Mode & OC Dir $\rightarrow$ Add & 0.623 [0.567, 0.678] \\
Mode & SL Add $\rightarrow$ Dir & 0.730 [0.684, 0.777] \\
Mode & SL Dir $\rightarrow$ Add & 0.714 [0.668, 0.761] \\
Wording & OC Add, T1 $\rightarrow$ T2--T10 & 0.560 [0.506, 0.614] \\
Wording & OC Dir, T1 $\rightarrow$ T2--T10 & 0.582 [0.532, 0.631] \\
Wording & SL Add, T1 $\rightarrow$ T2--T10 & 0.534 [0.494, 0.575] \\
Wording & SL Dir, T1 $\rightarrow$ T2--T10 & 0.586 [0.542, 0.630] \\
Backbone & OC $\rightarrow$ SL, Add & 0.642 [0.596, 0.689] \\
Backbone & OC $\rightarrow$ SL, Dir & 0.701 [0.655, 0.748] \\
Backbone & SL $\rightarrow$ OC, Add & 0.587 [0.525, 0.649] \\
Backbone & SL $\rightarrow$ OC, Dir & 0.693 [0.636, 0.748] \\
\bottomrule
\end{tabular}
\end{table}

Opportunity gains under same-condition selection/evaluation are 0.698 [0.650, 0.749], 0.752 [0.704, 0.800], 0.727 [0.682, 0.772], and 0.745 [0.700, 0.790]. Robust on-policy switch rates are 42.5\%, 46.6\%, 52.9\%, and 53.5\%. These are upper bounds because replan directly observes the evaluator's counterfactual utilities. Cross-condition cells instead select the order under one condition and evaluate it under another; their positivity is not guaranteed by construction. These evaluations preserve scenes and semantic targets, so they test robustness across evaluator changes rather than distribution shift. None of these analyses tests whether a selector can predict the choices cheaply or whether an encoder can execute the evidence groups independently.

\section{Role-balanced follow-up details}
\label{app:role-balanced}

The role-balanced follow-up contains 1,200 scenes and rotates the three abstract roles across all six assignments of color, shape, and texture. At $\epsilon=0.05$, intended-pair reversal rates are 68.37\% versus 0.88\% for OpenCLIP additive, 72.15\% versus 9.00\% for OpenCLIP direct render, 71.82\% versus 0.95\% for SigLIP additive, and 76.95\% versus 9.08\% for SigLIP direct render in positive versus negative constructions. The corresponding positive-minus-negative differences are 67.48, 63.15, 70.87, and 67.87 percentage points, respectively, with all four scene-bootstrap intervals excluding zero. All 24 permutation-specific backbone/mode contrasts are likewise positive with intervals above zero; the smallest lower bound is 46.4 percentage points.

The role-balanced result should not be described as attribute-role invariance. Seven unadjusted descriptive attribute-identity contrasts in positive regimes exclude zero. The most consistent difference is lower reversal when the initially strong attribute is color rather than texture: $-12.15$ pp [$-20.50$, $-3.85$] for OpenCLIP additive, $-12.85$ pp [$-20.30$, $-5.30$] for OpenCLIP direct render, and $-7.20$ pp [$-12.55$, $-1.85$] for SigLIP additive. These contrasts are descriptive and unadjusted for multiplicity; they establish residual attribute dependence without overturning the across-role coverage result.

\section{Prospective predictability follow-up}
\label{app:predictability}

The predictor follow-ups keep the role-balanced scene split fixed and exclude regime labels, role labels, future utilities, unacquired evidence, and other privileged information from inputs. Static-determinacy diagnostics first verify that the task is not trivial under legal static information. Exact static conditional entropy is 0.642, 0.631, 0.722, and 0.669 bits for OpenCLIP additive/direct and SigLIP additive/direct, respectively; 71.8--79.7\% of role-balanced rows belong to ambiguous exact-static strata, with 976--1,043 opposite-label matched pairs per condition.

A linear stateful predictor over the position-indexed acquired posterior does not obtain a reliable held-out advantage over no-history controls on role-balanced data. Replacing this representation with a candidate-permutation-invariant posterior representation and using shared action-conditioned linear, gradient-boosted-tree, and two-layer MLP predictors improves the older 800-scene benchmark in some conditions, but neither nonlinear model yields a significant matched-ambiguity cross-entropy improvement in any of the four role-balanced conditions. This rules out the simplest explanation that the role-balanced failure is specific to a linear classifier or raw candidate indexing.

The final predictor uses the acquired frozen visual embeddings directly. Each candidate is encoded by a small shared $64\rightarrow32$ network; pooled state features feed a shared $64\rightarrow32\rightarrow16\rightarrow1$ action scorer. Stateful, no-history, and shuffled models use identical parameter shapes, and hyperparameters are selected by validation cross-entropy without test access. The final matched-ambiguity stateful-minus-no-history cross-entropy improvements are shown in Table~\ref{tab:embedding-predictor}. Scene-pooled improvement is $-0.03120$ bits [95\% CI $-0.07280$, 0.00840]. The matched-subset pooled stateful-minus-no-history utility difference is 0.02174 [$-0.05271$, 0.09528], the all-test pooled utility difference is 0.00566 [$-0.01667$, 0.02875], and stateful-minus-shuffled utility on matched ambiguity is $-0.01744$ [$-0.07336$, 0.03906]. The predefined stop rule is therefore not met, and we do not continue to larger predictor architectures.

\end{document}